\documentclass[journal]{IEEEtran}
\newcommand{\equal}[1]{\underset{#1}{=}}
\newcommand{\equalo}[1]{\underset{#1}{\simeq}}
\usepackage{subcaption}

\usepackage{graphicx}
\usepackage{xcolor}
\usepackage{booktabs}
\usepackage{multirow}

\usepackage[hidelinks]{hyperref}
\usepackage{tikz}
\usetikzlibrary{arrows,positioning}
\usetikzlibrary{calc}
\usetikzlibrary{shapes}

\usepackage{amsmath,amssymb,amsfonts}
\usepackage{algorithm}
\usepackage{algpseudocode}
\usepackage{pgfplots}
\usepackage{pgfplotstable}

\pgfplotsset{compat = 1.18}

\newcommand{\virg}[1]{\textquotedblleft#1\textquotedblright}

\usepackage{csquotes}

\ifCLASSINFOpdf

\else

\fi

\begin{document}

\title{POMDP-Driven Cognitive Massive MIMO Radar: Joint Target Detection-Tracking In Unknown Disturbances}

\author{Imad~Bouhou,~\IEEEmembership{}
        Stefano~Fortunati,~\IEEEmembership{}
        Leila~Gharsalli,~\IEEEmembership{}
        Alexandre~Renaux.~\IEEEmembership{}
        
\thanks{Imad Bouhou is with Université Paris-Saclay, CNRS, CentraleSupélec, Laboratoire des signaux et systèmes, 91190, Gif-sur-Yvette, France \& DR2I-IPSA,
94200, Ivry-sur-Seine, France. (e-mail imad.bouhou@centralesupelec.fr).

Stefano Fortunati is with SAMOVAR, Télécom SudParis, Institut Polytechnique de Paris, 91120, Palaiseau, France (e-mail stefano.fortunati@telecom-sudparis.eu).

Leila Gharsalli is with DR2I-IPSA, 94200, Ivry-sur-Seine. (e-mail: leila.gharsalli@ipsa.fr).

Alexandre Renaux is with Université Paris-Saclay, CNRS, CentraleSupélec, Laboratoire des signaux et systèmes, 91190, Gif-sur-Yvette, France. (e-mail: alexandre.renaux@universite-paris-saclay.fr).
}
}

\markboth{}%
{Shell \MakeLowercase{\textit{et al.}}: Bare Demo of IEEEtran.cls for IEEE Journals}

\maketitle

\begin{abstract}
The joint detection and tracking of a moving target embedded in an unknown disturbance represents a key feature that motivates the development of the cognitive radar paradigm. Building upon recent advancements in robust target detection with multiple-input multiple-output (MIMO) radars, this work explores the application of a Partially Observable Markov Decision Process (POMDP) framework to enhance the tracking and detection tasks in a statistically unknown environment. In the POMDP setup, the radar system is considered as an intelligent agent that continuously senses the surrounding environment, optimizing its actions to maximize the probability of detection $(P_D)$ and improve the target position and velocity estimation, all this while keeping a constant probability of false alarm $(P_{FA})$. The proposed approach employs an online algorithm that does not require any apriori knowledge of the noise statistics, and it relies on a much more general observation model than the traditional range-azimuth-elevation model employed by conventional tracking algorithms. Simulation results clearly show substantial performance improvement of the POMDP-based algorithm compared to the State-Action-Reward-State-Action (SARSA)-based one that has been recently investigated in the context of massive MIMO (MMIMO) radar systems.
\end{abstract}


\IEEEpeerreviewmaketitle

\section{Introduction}
Cognitive radar, first introduced by Haykin in \cite{cr_haykin_way_of_future}, represents a major advancement in radar technology by incorporating cognitive processing capabilities. Unlike traditional systems, the cognitive radar continuously learns from its environment, dynamically optimizing its waveforms and operational parameters through real-time feedback between the transmitter and receiver. This intelligent feedback loop significantly enhances system adaptability and performance \cite{cr_def}.
Drawing inspiration from the biological perception-action cycle, cognitive radar systems actively perceive their environment, learning crucial information about the targets and background noise. This acquired knowledge enables intelligent waveform choices, as demonstrated in \cite{cr_waveform_design} or adaptive waveform selection using Bayesian approaches \cite{cr_waveform_adaptive}, resulting in a highly agile and responsive systems.

Some surveys \cite{cr_road_to_reality}, \cite{cr_overview} have examined cognitive radar's practical applications, challenges, and efficient decision-making capabilities in dynamic environments by linking it to machine learning, optimization, and Bayesian filtering approaches to make the transmitter fully cognitive and functional.

In this work, we consider a MMIMO radar whose goal is to detect and track a target embedded in a \textit{non-Gaussian disturbance with unknown statistical properties}. 
The fact that the radar is composed of a \virg{massive} number of antenna channels enables us to achieve remarkable robustness properties that allow us to obtain performance guarantees in a much more general scenario than the standard and unrealistic Gaussian case. Specifically, as shown in \cite{massive_mimo_wald}, the massive amount of virtual antenna channels allows for the derivation of a robust Wald-type detector able to guarantee a Constant False Alarm Rate (CFAR), even when the distribution of the noise is unknown, as long as its second-order moment decays at a polynomial rate.

However, despite its notable CFAR capacity, the robust Wald-type detector cannot automatically maximize the $P_D$, unlike Neyman-Pearson-based detectors, since the noise statistics remain unknown. To address this crucial aspect, the authors in \cite{rl_mimo_aya, Lisi} proposed a Reinforcement Learning (RL) strategy that maximizes the probability of detection $P_D$ using the SARSA algorithm to sequentially adjust the MMIMO waveform matrix without needing apriori knowledge of the disturbance statistics. While the algorithm proposed in \cite{rl_mimo_aya, Lisi} effectively detects weak targets in stationary or slow-moving scenarios, it experiences significant performance degradation in highly dynamic environments. In such cases, maximizing the $P_D$ is insufficient, and tracking capabilities must be incorporated into the MMIMO framework. 

Traditional radar tracking methods model disturbances with additive, zero-mean, Gaussian distributions (with possibly unknown covariance matrix) and use the traditional range-azimuth-elevation model as an observation model. This conventional framework enables the application of Bayesian filtering techniques, which were previously employed by cognitive radar frameworks in \cite{cr_haykin_way_of_future}, \cite{cr_filtering_kristine}, \cite{cr_scheduling} to estimate the current environmental state and predict its evolution. 

To overcome these two critical limitations, i.e., \textit{i)} the need for apriori knowledge of the disturbance statistics and \textit{ii)} the stationarity of the targets, this work proposes a cognitive radar framework that is based on a MMIMO radar system able to merge the robustness/CFAR property of the Wald-type detector \cite{massive_mimo_wald} with the ability of Reinforcement Learning (RL)-based algorithm to maximize the $P_D$ while enhancing tracking performance. Specifically, to address the possible non-stationarity of the scenario, the joint detection-tracking problem is modeled as a Partially Observable Markov Decision Process (POMDP) \cite{pomdp}. In simple terms, the \virg{observable part} of the process is the detection made by the robust Wald-type detector and the inherent estimation of the target power.  However, the \virg{unobservable part} --- that must instead be deduced from the observations --- is the target's actual state, specifically, its Cartesian coordinates in a given reference frame and the relevant velocity vector. 

POMDPs, as opposed to a Markov Decision Process (MDP), can handle sequential decision-making when the states are considered to be hidden. The agent when handling POMDPs, has to make actions based on the history of all the observations it received in the past. POMDPs are hard to solve, so approximation approaches are used, such as Deep recurrent Q-learning networks (DRQN) \cite{drqn} or online-learning algorithms that are based on the UCT algorithm \cite{uct} like Partially Observable Monte-Carlo Planing (POMCP) \cite{pomcp}.

In radar applications, some previous works have adopted the POMDP design to track targets. For instance, in \cite{cr_attention_drqn}, the authors proposed a DRQN equipped with an attention-head \cite{attention} in addition to a long-short term memory (LSTM) to be able to track a changing number of targets. In \cite{applying_dqn_1}, \cite{applying_dqn_2}, the authors applied Deep Q-learning Networks (DQN) \cite{dqn} and managed to improve tracking precision. However, both DQN and DRQN frameworks are affected by the same limitation: they require a representative dataset of the environment. This may be an unrealistic requirement since it would be difficult to collect a training dataset that contains examples for all the potential scenarios, given all the possible disturbance distributions and all the potential target trajectories. For this reason, it is preferable to use online algorithms that plan and adapt themselves sequentially to the information gathered in real-time.

Some previous works have already exploited online POMDP solvers to handle target tracking in radar applications; for instance, in \cite{cr_mcts_dpw}, the authors showed that the use of Monte Carlo Tree Search with double progressive widening (MCTS-DPW) \cite{mcts_dpw} improves tracking performance. A similar approach was proposed in \cite{cr_mcts_vpw} but with more guidance to how the tree search explores the action space, and this is done using Voronoi progressive widening (MCTS-VPW)\cite{mcts_vpw}.

In this work, we propose an original approach able to merge the POMCP machinery with the statistical robustness of the robust Wald-detector \cite{massive_mimo_wald} in order to come up with an algorithm providing nearly optimal tracking performance in the presence of \textit{unknown disturbance statistics}. The \virg{disturbance agnostic} capacity represents the crucial advantage of our proposed method with respect to competing algorithms, e.g., \cite{cr_mcts_vpw} and \cite{cr_mcts_dpw}, that require the apriori knowledge of the disturbance distribution. Further details on this fundamental point are detailed in \ref{sec:pomcp}.


\textbf{Notations}:
In this paper, matrices are denoted by uppercase letters $\mathbf{A}$ and vectors by lowercase letters $\mathbf{a}$. $(\cdot)^T$, $(\cdot)^H$, and $(\cdot)^*$ represent transpose, conjugate transpose, and complex conjugate, respectively. $\mathbf{I}_N$ is the $N \times N$ identity matrix and $\mathbf{0}_N$ is a zero vector of size $N$. $\otimes$ denotes Kronecker product. A closed intervals between $a$ and $b$ is denoted as $[a,b]$, and sets as $\{a,b\}$. $|\cdot|$ is the absolute value. For probability distributions, the chi-squared distribution with $k$ degrees of freedom is denoted as $\chi^{2}_k\left( \delta \right)$, where $\delta$ is the non-centrality parameter (then the distribution is central if $\delta=0$). The real Gaussian distribution with mean $\mu$ and a variance $\sigma^{2}$ is represented as $\mathcal{N}\left( \mu,\sigma^{2} \right)$ (and $\mathcal{CN}\left( \mu,\sigma^{2} \right)$ when it is a circular complex Gaussian). Finally, the exponential distribution with a rate parameter $x$ is written as $\text{Exp}(x)$.

\section{PROBLEM FORMULATION}
\label{sec:problem_formulation}
This section presents the radar system model, defines mathematically what is a POMDP, and summarizes the POMCP
algorithm employed in this paper.

\subsection{System Model}
The MMIMO signal model considered here is the same as the one used in \cite{massive_mimo_wald} and \cite{rl_mimo_aya}. To avoid redundancy, in the following, we recall only the main points that are needed for the subsequent derivations while we refer to  \cite{massive_mimo_wald} and \cite{rl_mimo_aya} for an in-depth description. 

We consider a co-located MIMO radar equipped with $N_T$ transmit and $N_R$ receive physical antennas. Following the standard MIMO theory, we indicate as $N = N_T N_R$ the total number of \textit{virtual spatial antenna channels}. The radar field of view is assumed to be discretized into $L$ angle bins $\{\theta_l; l=1,.., L \}$, and in total the system transmits $T_{\text{max}}$ scans, $t \in \{0,.., T_{\text{max}} - 1\}$. Therefore, the detection problem can be re-formulated for an angle bin $l$ and a time step $t$ as \cite{rl_mimo_aya, massive_mimo_wald}:
\begin{equation}
    \begin{split}
        H_0 &: \mathbf{y}_{t+1,l} = \mathbf{c}_{t+1,l},\\
        H_1 &: \mathbf{y}_{t+1,l} = \alpha_{t+1,l}\mathbf{v}_{t,l} + \mathbf{c}_{t+1,l},
    \end{split}
    \label{h0_h1}
\end{equation}
where $\mathbf{c}_{t+1,l} \in \mathbb{C}^{N}$ is a random vector of unknown probability density function $p_C$, representing the disturbance. We assume that the $N$ entries of the disturbance vector are sampled from a circular complex random process $\{c_{t+1,l,n}, \forall n \}$. We only assume that its auto-correlation function exists and decays \textit{at least} at a polynomial rate \cite[Assumption 1]{massive_mimo_wald}. The unknown deterministic scalar $\alpha_{t+1,l} \in \mathbb{C}$ stands for the radar-cross section and the two-way path loss. The known vector $\mathbf{v}_{t,l} \in \mathbb{C}^{N}$ is defined as in \cite{massive_mimo_wald} and \cite{rl_mimo_aya}:
\begin{equation}
    \mathbf{v}_{t,l} = (\mathbf{W}_t^T \mathbf{a}_T(\theta_l))\otimes\mathbf{a}_R(\theta_l),
    \label{h_w}
\end{equation}
where $\mathbf{W}_t \in \mathbb{C}^{N_T \times N_T}$ is the waveform matrix and $\mathbf{a}_T(\theta_l)$ and $\mathbf{a}_R(\theta_l)$ are the transmit and receive steering vectors, assumed to be perfectly known. The waveform matrix $\mathbf{W}_t$ that focuses the transmitted energy on angle bin $\theta_l$ is determined by solving the following optimization problem \cite{waveform_design, EuRAD}:
\begin{equation}
\begin{split}
    \text{max}_{\mathbf{W}} \ \  & \mathbf{a}_T^T(\theta_l) \mathbf{W}\mathbf{W}^H\mathbf{a}_T^*(\theta_l) \\
    \text{s.t.} \ \ & \text{Tr}(\mathbf{W}\mathbf{W}^H) = P_T,
\end{split}
\label{w_problem}
\end{equation}
where $P_T$ is the total transmit power of the radar. \\
Clearly, the waveform matrix $\mathbf{W}_t$ is a solution to \eqref{w_problem} if and only if it is a square-root of the matrix $\frac{P_T}{N_T}\mathbf{a}_T^*(\theta_l)\mathbf{a}^T_T(\theta_l)$.

As shown in \cite{massive_mimo_wald}, the hypothesis testing problem in \eqref{h0_h1} can be solved using the following robust Wald-type test:

\begin{equation}
    \Lambda_{t+1,l} = 2 |\hat{\alpha}_{t+1,l}|^2 \frac{||\mathbf{v}_{t,l}||^4}{\mathbf{v}_{t,l}^H \widehat{\mathbf{\Sigma}}_{t+1,l} \mathbf{v}_{t,l} } \underset{H_0}{\overset{H_1}{\gtrless}} \lambda,
\end{equation}
where $\widehat{\mathbf{\Sigma}}_{t+1,l}$ is an estimator of the disturbance covariance matrix that can be computed the same way as in \cite{massive_mimo_wald}, \cite{erratum_massive_mimo}, while $\hat{\alpha}_{t+1,l} = (\mathbf{v}_{t,l}^H \mathbf{y}_{t+1,l})/||\mathbf{v}_{t,l}||^2$ is an estimator of the parameter $\alpha_{t+1,l}$. The threshold $\lambda$ is chosen to keep the probability of false alarm below a predetermined value, say $P_{FA}$, by solving the following equation:
\begin{equation}
    P_{FA}= \mathrm{Pr}\{\Lambda_{t+1,l} > \lambda | H_0\}.
    \label{p_fa_equation}
\end{equation}
In \cite{massive_mimo_wald}, the authors proved that $\Lambda_{t+1,l}|H_0 \underset{N \to \infty}{\sim} \chi_2^2(0)$. Therefore, \eqref{p_fa_equation} becomes:
\begin{equation}
    P_{FA}\equal{N\to+\infty}\text{exp}(- \lambda /2).
\label{p_fa_equation_1}
\end{equation}
The threshold $\lambda$ is computed by choosing a desired $P_{FA}$ using $\lambda = -2 \text{log}(P_{FA})$. \\
The probability of detection $P_D$ is defined by the equation:
\begin{equation}
    P_{D} = \mathrm{Pr}\{\Lambda_{t+1,l} > \lambda | H_1\}.
    \label{p_d}
\end{equation}
In \cite{massive_mimo_wald}, the authors proved that $\Lambda_{t+1,l}|H_1 \underset{N \to \infty}{\sim} \chi_2^2(\zeta_{t+1,l})$ where $\zeta_{t+1,l} = 2 |\alpha_{t+1,l}|^2 ||\mathbf{v}_{t,l}||^4/(\mathbf{v}_{t,l}^H \mathbf{\Sigma} \mathbf{v}_{t,l} )$ and $\mathbf{\Sigma}$ is the true unknown disturbance covariance. Therefore, \eqref{p_d} becomes:
\begin{equation}
    P_{D} \equal{N\to+\infty} Q_1(\sqrt{\zeta_{t+1,l}}, \sqrt{\lambda}),
    \label{p_d_0}
\end{equation}
where $Q_1$ is the Marcum Q function of order 1. \\
Given that the probability density function of the disturbance $p_C$ is unknown, $\zeta_{t+1,l}$ needs to be estimated by $\Lambda_{t+1,l}$. Hence, the probability of detection can be approximated by:
\begin{equation}
    P_{D} \equalo{N\to+\infty} Q_1 \left(\sqrt{2 |\hat{\alpha}_{t+1,l}|^2 \frac{||\mathbf{v}_{t,l}||^4}{\mathbf{v}_{t,l}^H \widehat{\mathbf{\Sigma}}_{t+1,l} \mathbf{v}_{t,l} }}, \sqrt{\lambda} \right).
    \label{p_d_1}
\end{equation}
The waveform matrix $\mathbf{W}_t$ has to be chosen to maximize the probability of detection $P_D$.

\subsection{Reinforcement learning (POMDPs): a recall}

Reinforcement learning (RL) is a branch of machine learning in which an agent learns to make decisions by interacting with an environment \cite{rl_sutton}. The main goal is to maximize a cumulative reward over time. In this framework, the agent explores various actions, observes their consequences, and adjusts its strategy accordingly.

A POMDP \cite{pomdp} can be described by the tuple $(\mathcal{S}, \mathcal{O}, \mathcal{A}, \mathcal{P}, \Omega, \mathcal{R})$, where 
$\mathcal{S}$ represents the set of states, 
$\mathcal{O}$ represents the set of observations,
$\mathcal{A}$ the set of available actions. 
The state transition probabilities
\begin{equation}
    \mathcal{P}_{s,s'}^{a} = p(S_{t+1} = s'|S_t = s, A_t = a),
\end{equation}
define the likelihood of transitioning from the current state $s$ to a new state $s'$ after taking action $a$. Similarly, the observation probabilities
\begin{equation}
\Omega_{s', o}^{a} = p(O_{t+1} = o |S_{t+1} = s', A_t = a), 
\end{equation}
specify the likelihood of observing $o$ after executing action $a$ and arriving at state $s'$. Finally, $\mathcal{R}_{s,s'}^a$ is the reward obtained after executing action $a$, being in a state $s$ and reaching state $s'$. Additionally, one needs to define other quantities useful for the following. For instance, a history $h_t$ is defined by:
\begin{equation}
    h_t = (a_0, o_1, ..., a_{t-1}, o_t), 
\end{equation}
which is a sequence of the previously taken actions and the obtained observations. The agent's goal is to develop a policy
\begin{equation}
    \pi(a|h) = p(A_t = a |H_t = h),
\end{equation}
that maps histories to a probability
distribution over actions. In addition to the history, the agent can also build a posterior distribution: 
\begin{equation}
b(s|h) = p(S_t=s|H_t=h),
\end{equation} called the belief state. \\
In the POMDP, the value function $V_{\pi}(h)$ is the expected return starting from a history $h$ and following a policy $\pi$,
\begin{equation}
    V_{\pi}(h) = \mathbb{E} \left[\sum_{t=0}^{\infty} \gamma^t \mathcal{R}_{s_t,s_{t+1}}^{a_t} \ \vline \
    \begin{aligned}
    & h_0 = h; s_0 \sim b(.|h_0) \\
    &  t \geq 0: a_t \sim \pi(.|h_t) \\ 
    &  t \geq 0: s_{t+1} \sim \mathcal{P}_{s_t}^{a_t} \\
    &  t \geq 0: o_{t+1} \sim \Omega_{s_{t+1}}^{a_t} \\
    & h_{t+1} = (h_t, a_t, o_{t+1})
\end{aligned} \right],
\end{equation}
where $\gamma \in (0,1) $ is called the discount factor.
The optimal value function $V^*(h) = \max_{\pi} V(h)$ is the value function computed with the optimal policy $\pi^*$.\\
When the agent has executed an action $a$ and observed an observation $o$, it has to update the belief state with the new information $(a,o)$ using the following equation:
\begin{equation}
    b(s'|h,a,o) \ \propto \ \Omega_{s', o}^{a} \sum\nolimits_{s} \mathcal{P}_{s, s'}^{a} b(s|h).
    \label{belief_state_update}
\end{equation}

\subsection{POMCP algorithm}
\label{sec:pomcp}
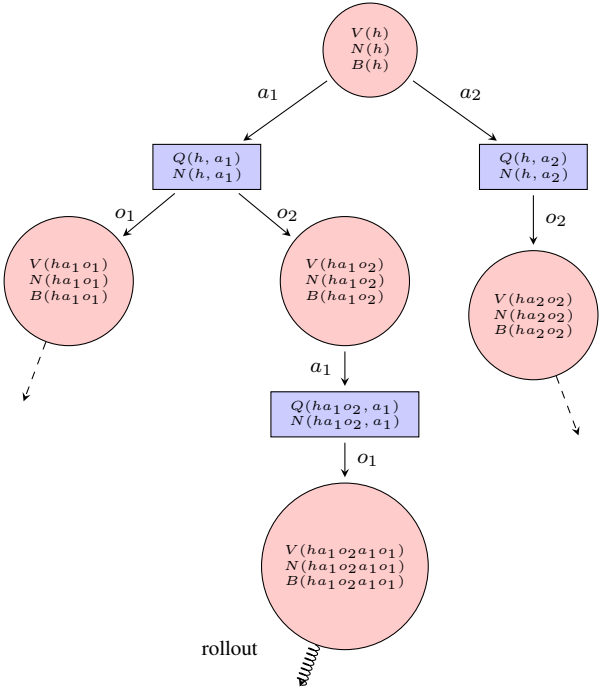
\begin{figure}
\centering
\begin{tikzpicture}[
    >=stealth,
    every node/.style={ellipse, draw, minimum width=1.2cm, minimum height=0.8cm, inner sep=1pt, fill=red!20},
    node distance=0.8cm and 1cm,
    square/.style={rectangle, draw, minimum size=0.6cm, fill=blue!20},
    myarrow/.style={->, shorten >=2pt, shorten <=2pt},
    label/.style={draw=none, font=\footnotesize, fill=none}
]
\node (h) {\tiny\begin{tabular}{c}$V(h)$ \\ $N(h)$ \\ $B(h)$ \end{tabular}};
\node[square] (ha1) [below left=of h] {\tiny\begin{tabular}{c}$Q(h,a_1)$ \\ $N(h,a_1)$\end{tabular}};
\node[square] (ha2) [below right=of h] {\tiny\begin{tabular}{c}$Q(h,a_2)$ \\ $N(h,a_2)$\end{tabular}};
\node (ha1o1) [below left=0.6cm and 0.5cm of ha1] {\tiny\begin{tabular}{c}$V(ha_1o_1)$ \\ $N(ha_1o_1)$ \\ $B(ha_1o_1)$ \end{tabular}};
\node (ha1o2) [below right=0.6cm and 0.5cm of ha1] {\tiny\begin{tabular}{c}$V(ha_1o_2)$ \\ $N(ha_1o_2)$ \\ $B(ha_1o_2)$ \end{tabular}};
\node (ha2o2) [below=0.8cm of ha2] {\tiny\begin{tabular}{c}$V(ha_2o_2)$ \\ $N(ha_2o_2)$ \\ $B(ha_2o_2)$\end{tabular}};
\node[square] (ha1o2a1) [below=0.6cm of ha1o2] {\tiny\begin{tabular}{c}$Q(ha_1o_2,a_1)$ \\ $N(ha_1o_2,a_1)$\end{tabular}};
\node (ha1o2a1o1) [below=0.6cm of ha1o2a1] {\tiny\begin{tabular}{c}$V(ha_1o_2a_1o_1)$ \\ $N(ha_1o_2a_1o_1)$ \\ $B(ha_1o_2a_1o_1)$ \end{tabular}};
\draw[myarrow] (h) -- node[label, midway, above left] {$a_1$} (ha1);
\draw[myarrow] (h) -- node[label, midway, above right] {$a_2$} (ha2);
\draw[myarrow] (ha1) -- node[label, midway, left] {$o_1$} (ha1o1);
\draw[myarrow] (ha1) -- node[label, midway, right] {$o_2$} (ha1o2);
\draw[myarrow] (ha2) -- node[label, midway, right] {$o_2$} (ha2o2);
\draw[myarrow] (ha1o2) -- node[label, midway, left] {$a_1$} (ha1o2a1);
\draw[myarrow] (ha1o2a1) -- node[label, midway, right] {$o_1$} (ha1o2a1o1);
\draw[dashed,->] (ha1o1) -- +(-0.6cm,-1.6cm);
\draw[dashed,->] (ha2o2) -- +(0.6cm,-1.6cm);
\draw[decoration={coil,aspect=0.3,segment length=2pt,amplitude=2pt},decorate,->] 
    (ha1o2a1o1) -- +(-0.6cm,-1.6cm);
\node[label, left] at ($(ha1o2a1o1.south west) + (-0.3,-0.3)$) {rollout};
\end{tikzpicture}
\caption{POMCP Tree illustration with two actions and two observations.}
\label{pomcp_illustration}
\end{figure}

The POMCP algorithm \cite{pomcp} is an online planning algorithm in large POMDPs when both the actions and observations are discrete. It is an extension of the UCT algorithm \cite{uct} to POMDPs by building a MDP whose states are defined by histories. \\
The POMCP needs a black-box generator $\mathcal{G}(s, a) = (s', o, r)$ where $r = \mathcal{R}_{s,s'}^a$, instead of knowing the distributions $\mathcal{P}$ and $\Omega$ explicitly.
The POMCP tree is initialized at the root with a belief set $B$, an estimation of the belief state $b(s|h)$, containing $N_p$ unweighted particles. \\
In Fig. \ref{pomcp_illustration}, each red node in the tree contains $(V(h), N(h), B(h))$, representing the value of the history $h$, the number of times $h$ has been visited, and the belief set, respectively. The red nodes $(V(h), N(h), B(h))$ have blue children nodes defined by $(Q(h, a), N(h, a))$ for all possible actions $a \in \mathcal{A}$. Here, $N(h, a)$ is the number of times where action $a$ was taken in history $h$, and $Q(h, a)$ is the value of being in history $h$ followed by the execution of the action $a$. Blue nodes generate an observation $o$, leading to new red child nodes. These new nodes are characterized by an updated history combining the previous history $h$ with the new information pair $(a,o)$.

The agent runs $N_{\text{sim}}$ simulations through the tree search and computes the value of the current history $V(h)$ at the root of the tree. Algorithm \ref{alg:simulate} presents the simulation process. It starts by sampling a state from the belief set $B$ at the root of the tree and then selecting the actions that maximize the Upper Confidence Bound (UCB1) \cite{ucb1} criterion given the actions of the child nodes  $Q^{UCT}(h, a) = Q(h, a) + c \sqrt{\frac{\log(N(h))}{N(h, a)}}$ where $c$ is a hyperparameter set to balance between exploration and exploitation. When a leaf is encountered, a new node is added, from which the rollout stage begins, which consists of running a simulation starting from the newly added node using some rollout policy $\pi_\text{{rollout}}$.
To run the simulation through the tree, the algorithm uses a black-box generator $\mathcal{G}(s, a)$.
\begin{algorithm}
\caption{Simulate}
\label{alg:simulate}
\begin{algorithmic}[1]
\Procedure{\texttt{SIMULATE}}{$s$, $h$, $depth$}
    \If{$\gamma^{depth} < \varepsilon$}
        \State \Return 0
    \EndIf
    \State $a \gets \arg\max_b Q(h, b) + c \sqrt{\frac{\log N(h)}{N(h,b)}}$
    \State $(s', o, r) \sim \mathcal{G}(s, a)$
    \If{$\texttt{Node}(h,a,o) \notin \texttt{Tree}$} 
        \State Add $\texttt{Node}(h,a,o)$
        \State \Return \texttt{ROLLOUT}$(s', h, depth)$
    \EndIf

    \State $R \gets r + \gamma \cdot \texttt{SIMULATE}(s', hao, depth + 1)$
    \If{$depth \neq 0$}
        \State $B(h) \gets B(h) \cup \{s\}$
    \EndIf
    \State $N(h) \gets N(h) + 1$
    \State $N(h,a) \gets N(h,a) + 1$
    \State $Q(h,a) \gets Q(h,a) + \frac{R - Q(h,a)}{N(h,a)}$
    \State \Return R
\EndProcedure

\Procedure{ \texttt{ROLLOUT}}{$s$, $h$, $depth$}
    \If{$\gamma^{depth} < \varepsilon$}
        \State \Return 0
    \EndIf
    \State $a \sim \pi_{\text{rollout}}(.|h) $
    \State $(s', o, r) \sim \mathcal{G}(s, a)$
    \State \Return $r + \gamma \texttt{ROLLOUT}(s', hao, depth + 1)$
\EndProcedure
\end{algorithmic}
\end{algorithm}

\begin{algorithm}
\caption{Solve}
\label{alg:solve}
\begin{algorithmic}[1]
\Require $N_{\text{sim}}$ \Comment{Number of simulations.}
\Require $B$ \Comment{Belief set at the root.}
\Procedure{\texttt{POMCP}}{$h$}
    \For{each simulation $i=1,..,N_{\text{sim}}$}
        \State $s \sim B$
        \State  \texttt{SIMULATE}$(s, h, 0)$
    \EndFor
    \State \Return $\arg\max_a Q(h, a)$
\EndProcedure
\end{algorithmic}
\end{algorithm}
As stated in Theorem 1 in \cite{pomcp}, as the number of simulations increases, the value function $V(h)$ computed by the POMCP at the root of the tree converges in probability to $V^*(h)$. \\
At the end of the simulations, the agent chooses the action $a^* = \arg\max_{a \in \mathcal{A}} Q(h, a)$, as shown in Algorithm \ref{alg:solve}. \\
After executing the optimal action and observing $o$, the belief update proceeds as follows: the algorithm selects a random particle $s$ from $B$, generates $(s', o', r)$ using $\mathcal{G}(s, a)$, and compares $o'$ to $o$. If $o'=o$, then $s'$ is added to the new belief set $B'$. This process is repeated until the new belief $B'$ contains $N_p$ particles.

\section{Cognitive radar as a POMDP}
\label{sec:pomdp_components}

In this section, we explain how the aforementioned POMDP definitions can be fitted into the radar framework and provide a full algorithm description. In this work, the algorithm is used with the assumption of having only one single target in the environment.

\subsection{Action space}
The action is related to the set of possible waveform matrices from which the radar can choose. The radar's angular resolution is represented by $L$ angle bins, so the radar will have $L$ actions, which means, the radar can choose one waveform matrix out of $L$ that allows it to focus all the energy in a single angle bin. At time step $t$, the radar simply chooses an angle bin $l \in \{1,2 ... L\}$ associated with an angle $\theta_l$, and the waveform matrix $\mathbf{W}_t$ is computed as the solution of the optimization problem (\ref{w_problem}), i.e., the matrix $\mathbf{W}_t$ is the square root of the matrix $\frac{P_T}{N_T}\mathbf{a}_T^*(\theta_l)\mathbf{a}^T_T(\theta_l)$.

\subsection{State space}
The state space consists of the target's possible positions and velocities. The state space consists of the target's possible positions and velocities.  At time step $t$, the state is defined as 
$\mathbf{s}_t=[x_t, V_{x,t}, y_t, V_{y,t}]^T$ where $[x_t, y_t]$ and $[V_{x,t}, V_{y,t}]$ denote the target position and velocity vectors respectively. \\
The following equation describes the dynamics of the target:
\begin{equation}
    \mathbf{s}_{t+1} = \mathbf{A}\mathbf{s}_t + \mathbf{G}\mathbf{w}_t,
    \label{motion_model}
\end{equation}
where \(\mathbf{A}\) is the state transition block-matrix :
\begin{equation}
\mathbf{A} =  
\begin{bmatrix}
\mathbf{A}_b & \mathbf{0}_{2 \times 2} \\
\mathbf{0}_{2 \times 2} & \mathbf{A}_b \\
\end{bmatrix},
\mathbf{A}_b = 
\begin{bmatrix}
1 & \Delta t \\
0 & 1
\end{bmatrix}.
\end{equation}
The term \(\mathbf{G}\mathbf{w}_t\) represents the noise, and the matrix \(\mathbf{G}\) can also be written in block form as:
\begin{equation}
\mathbf{G} = 
\begin{bmatrix}
\mathbf{G}_b & \mathbf{0}_{2 \times 1} \\
\mathbf{0}_{2 \times 1} & \mathbf{G}_b \\
\end{bmatrix},
\mathbf{G}_b = 
\begin{bmatrix}
\Delta t^2 / 2 \\
\Delta t
\end{bmatrix}.
\end{equation}

\begin{equation}
    \mathbf{w}_t \sim \mathcal{N} \left( \mathbf{0}_2 , \sigma_s^2 \mathbf{I}_2 \right),
\end{equation}
where $\sigma_s$ is the standard deviation of the process noise. \\
Given \eqref{motion_model}, the future state only depends on the current state, i.e., in this case, the transition probability does not depend on the action taken by the radar.

\subsection{Observation space}
At time step $t$, the radar chooses an action $a$ associated with an angle bin $l$ from which a vector $\mathbf{v}_{t,l}$ is computed, then the radar observes the estimation of the parameter $|\alpha_{t+1, l}|$ when there is a detection.
\begin{equation}
o_{t+1} =
\begin{cases} 
|\hat{\alpha}_{t+1,l}| & \text{if} \ \ \Lambda_{t+1,l} \geq 
\lambda, \\
\emptyset & \text{otherwise},
\end{cases}
\label{alpha_obs}
\end{equation}

In accordance with the classical \virg{radar equation} theory, the parameter $|\alpha_{t+1,l}|$ exhibits an inverse proportional relationship with $R_{t+1}^2$ \cite{range_alpha}, where $R_{t+1}$ represents the range between the target and radar.
\begin{equation}
|\alpha_{t+1,l}| \ \propto \ 1/R_{t+1}^2.
\label{alpha_range}
\end{equation}
In \cite{massive_mimo_wald}, it has been shown that the estimated parameter $\hat{\alpha}_{t+1,l}$ is asymptotically distributed as:
\begin{equation}
    \frac{\hat{\alpha}_{t+1,l} - \alpha_{t+1,l}}{\widehat{\sigma}_{t,l}} \underset{N \to \infty}{\sim} \mathcal{CN}(0,1),
    \label{alpha_asymptotic}
\end{equation}
where $\widehat{\sigma}_{t,l} = \sqrt{\mathbf{v}_{t,l}^H \widehat{\mathbf{\Sigma}}_{t+1,l} \mathbf{v}_{t,l}}/ ||\mathbf{v}_{t,l}||^2$ is given in \cite{massive_mimo_wald}.\\
The POMCP is an online POMDP solver when the observations are discrete. Since, in our case, according to \eqref{alpha_obs}, the observation space is continuous, it must be transformed into a discrete one by choosing a discretization step $\beta_l$. \\
The result in \eqref{alpha_asymptotic} allows to approximate the distribution of \\ $|\hat{\alpha}_{t+1,l} - \alpha_{t+1,l}|^2$ with the exponential distribution $\text{Exp}(\widehat{\sigma}_{t,l}^2)$ when $N$ goes to infinity. Then, we define the discretization step $\beta_l$ as the real number verifying the following condition:
\begin{equation}
    \mathrm{Pr}\{ \big| \ |\hat{\alpha}_{t+1,l}|  - |\alpha_{t+1,l}| \ \big| < \beta_l \} \geq 0.95.
    \label{stepsize_choice}
\end{equation}
It is enough that $\beta_l$ verifies the following condition:
\begin{equation}
    \mathrm{Pr}\{ \big| \ \hat{\alpha}_{t+1,l}  - \alpha_{t+1,l} \ \big|^2 < \beta_l^2 \} = 0.95,
    \label{exp_step}
\end{equation}
to also verify the condition \eqref{stepsize_choice}.
Therefore, using the analytic form of the cumulative distribution function of the exponential distribution $\text{Exp}(\widehat{\sigma}_{t,l}^2)$, one gets $\beta_l = \sqrt{3} \widehat{\sigma}_{t,l}$.

\subsection{Reward function}
The reward function should encourage the radar to detect and track the target in the environment. In the POMDP definition, the reward function depends on the current state $\mathbf{s}$, the taken action $a$, and the next state $\mathbf{s}'$. The action $a$ is about choosing an angle bin $\theta_a$ where the target will be located in the future. \\
Let us indicate by $\theta_{\mathbf{s}'}$ the true future angle bin of the target. To encourage accurate target position prediction, the reward function is chosen as:
\begin{equation}
    \mathcal{R}_{s,s'}^a = \mathbf{1}\{\theta_a = \theta_{\mathbf{s}'}\}.
    \label{reward}
\end{equation}
Note that the reward function here does not depend on the current state $\mathbf{s}$.

\subsection{Simulation model}
As previously stated, the POMCP algorithm \cite{pomcp} needs a black-box generator $\mathcal{G}(s, a) = (s', o, r)$ to be able to run simulations through the tree search.\\ In the context of this paper, the noise disturbance $p_C$ is assumed to be unknown, which makes the observation probabilities unknown and consequently makes the POMCP impossible to use. \\
Fortunately, using the asymptotic distribution of the estimated parameter $\hat{\alpha}_{t+1,l}$, a generator can be put in place to be used by the POMCP and the particle filter to maintain the detection and tracking of a target in the environment.

\begin{algorithm}
\caption{Generator $\mathcal{G}(\mathbf{s}_t, a_t)$.}
\label{generator}
\begin{algorithmic}[1]
\Require $\mathbf{s}_t=(x_t, V_{x,t}, y_t, V_{y,t})^T$ and action $a_t$.
\Require $(\widehat{\sigma}_l)_{l \in \{1,..,L\}}$.
\State $\mathbf{s}_{t+1} \gets \mathbf{A}\mathbf{s}_t + \mathbf{G}\mathbf{w}_t$
\State $\theta_{t+1} \gets \texttt{GetAngleBin}(\mathbf{s}_{t+1})$
\State $l_{t} \gets \texttt{GetAngleBin}(a_t)$
\State $\alpha_{t+1} \gets \texttt{GetRCS}(\mathbf{s}_{t+1})$ \Comment{\eqref{alpha_range}}
\State $\widehat{\alpha}_{t+1} \gets \mathcal{CN}(\alpha_{t+1}, \widehat{\sigma}_{l_t}^2)$ \Comment{\eqref{alpha_asymptotic}}
\State $\Lambda_{t} \gets 2 |\widehat{\alpha}_{t+1}|^2/ \widehat{\sigma}_{l_t}^2$ \Comment{\eqref{lambda_reform}}

\If{$ l_{t} \neq \theta_{t+1}$} $o_{t+1} \gets \emptyset$
\ElsIf{$ l_{t} = \theta_{t+1}$}
    \If{$\Lambda_{t} \geq \lambda$} $o_{t+1} \gets |\hat{\alpha}_{t+1}| $
    \ElsIf{$\Lambda_{t} < \lambda$}
        \State $o_{t+1} \gets \emptyset$
    \EndIf
\EndIf
\State $r_{t} \gets \mathbf{1}\{l_t = \theta_{t+1} \}$
\State \Return $(s_{t+1}, o_{t+1}, r_{t})$
\end{algorithmic}
\end{algorithm}
In the Algorithm \ref{generator}, the \texttt{GetAngleBin} function determines the angle bin based on the target's coordinates or the radar's action. The \texttt{GetRCS} function computes the parameter $\alpha_{t+1,l} = |\alpha_{t+1,l}| e^{j\phi}$, where $\phi$ is uniformly sampled from $(0, 2\pi)$. According to \eqref{alpha_range}, the magnitude $|\alpha_{t+1,l}|$ depends on the target's position. This proposed generator requires parameters $(\widehat{\sigma}_l)_{l \in \{1,..,L\}}$ to be used in the asymptotic relation (\ref{alpha_asymptotic}). To simulate the detection, the decision statistic can be reformulated using $\widehat{\sigma}_{t,l}^2$ to obtain the following equation:
\begin{equation}
    \Lambda_{t+1,l} = 2 |\hat{\alpha}_{t+1,l}|^2/\widehat{\sigma}_{t,l}^2.
    \label{lambda_reform}
\end{equation}
The decision statistic inside the generator will take the same format as in \eqref{lambda_reform} by replacing $\widehat{\sigma}_{t,l}$ with $\widehat{\sigma}_{l}$.
\\
The generator will be used by the POMCP to run simulations in the tree search and will also be used by the unweighted particle filter to make predictions and fill up the belief set.

\section{Cognitive Radar Design}
\label{sec:cr_design}

This section introduces an unweighted particle filter capable of handling distribution agnostic observation models and then explains how the POMCP is incorporated into the cognitive radar framework. \\
Unlike standard tracking tasks with known observation models, our scenario involves an unknown observation model due to the unspecified disturbance $p_C$. To address this, the typical particle filter is replaced with an unweighted version. Additionally, note that it is employed with a non-standard observation model, departing from the conventional range-azimuth-elevation format used in most tracking applications.

\subsection{The particle filter}
The particle filter presented in this section is exactly the same as the one used by the POMCP to estimate the posterior. Here, instead of running simulations inside the tree search, this approach only makes a prediction about the target's position, and given that prediction, an action is selected.
\\
At time step $t$, an approximation of $b(.|h_t)$ is defined by the set $B_t$. Similarly to the POMCP, which is driven with rewards, the particle filter needs to predict the future hidden state of the target, hence, compute $\mathbb{E}( \mathbf{s}_{t+1}|h_t)$. \\
By definition of the target dynamics, the transition probabilities only depend on the previous state, therefore, $\mathcal{P}_{\mathbf{s},\mathbf{s}'}^a = \mathcal{P}_{\mathbf{s},\mathbf{s}'}$.
\begin{align}
    \mathbb{E}(\mathbf{s}_{t+1}|h_t) &= \int_{\mathbf{s}'} \mathbf{s}' p(\mathbf{s}_{t+1} = \mathbf{s}'| H_t = h_t) d\mathbf{s}' \\
    &= \int_{\mathbf{s}'} \mathbf{s}' \int_{\mathbf{s}} \mathcal{P}_{\mathbf{s},\mathbf{s}'}
    b(\mathbf{s}|h_t) d\mathbf{s} d\mathbf{s}' \\
    &= \int_{s} \mathbb{E}(\mathbf{s}_{t+1}|\mathbf{s}_t=\mathbf{s})b(\mathbf{s}|h_t) d\mathbf{s} \\
    & = \mathbb{E}_{\mathbf{s}_t \sim b(.|h_t)} \Big[ \mathbb{E}(\mathbf{s}_{t+1}|\mathbf{s}_t)\Big] \\
    &\approx \frac{1}{|B_t|} \sum_{\mathbf{s} \in B_t} \mathbb{E}(\mathbf{s}_{t+1}|\mathbf{s}_t=\mathbf{s})
\end{align}

In essence, when the particle filter computes the prediction, it predicts the angle bin where the target will be in the future, therefore it makes sense to select the action related to the predicted angle bin.

\subsection{POMCP in the cognitive radar setup}
\label{pomcp_in_cr_setup}
The radar chooses the orthogonal waveform matrix $\mathbf{W}_{\text{orth}} = \sqrt{\frac{P_T}{N_T}}\mathbf{I}_{N_T{}}$ at the beginning until there is a detection, then it uses the observations to estimate the coordinates of the target and initializes the velocities uniformly using an interval $[-V_{\text{max}}, V_{\text{max}}]$, where $V_{\text{max}}$ is some maximum velocity. During this phase, the standard deviations $(\widehat{\sigma}_l)_{l \in \{1,..,L\}}$ to use in the asymptotic relation \eqref{alpha_asymptotic} can be computed and saved to be used later in the generator. \\
In the Algorithm \ref{cr_design}, the full cognitive radar design is presented, and, particularly, how the POMCP is integrated into it.
\begin{algorithm}
\caption{Cognitive radar design.}
\label{cr_design}
\begin{algorithmic}[1]
\Require $N_{\text{sim}}$ \Comment{Number of simulations.}
\Require $B_0$ \Comment{Initial Belief set.}
\Require $\mathcal{G}$ \Comment{A black-box generator.}
\Require $(\widehat{\sigma}_l)_{l \in \{1,..,L\}}$ \Comment{Parameters for the generator.}
\Require $\beta_l = \sqrt{3} \widehat{\sigma}_l \ \ l=1,\cdots,L$  \Comment{Parameters for the generator.}
\For{each time step $t=0,..,T_{\text{max}}-1$}
    \State $a_t \gets \texttt{POMCP.Solve}(N_{\text{sim}}, B_t)$. \Comment{Algorithm \ref{alg:solve}}
    \State $l \gets \texttt{GetAngleBin}(a_t)$.
    \State Compute $\mathbf{W}_t$ from $l$.
    \State Receive the signal $\mathbf{y}_{t+1,l}$ and compute observations.
    \State Observe $o_{t+1}$ from \eqref{alpha_obs}.
    \State $B_{t+1} \gets \texttt{UpdateBelief}(B_t, a_t, o_{t+1})$.
\EndFor
\end{algorithmic}
\end{algorithm}

\section{Simulations}
\label{sec:simulations}

The disturbance model used here will be the same as in \cite{rl_mimo_aya} and is based on an auto-regressive process (AR) of order $p$.

\begin{equation}
    c_n = \sum\nolimits_{i=1}^{p} \rho_i c_{n-i} + w_n, \quad n \in (-\infty, +\infty),
\end{equation}
The process is driven by identically independent distributed i.i.d. $t$-distributed innovations $w_n$ with a probability density function $p_w$ defined as:

\begin{equation}
    p_w(w_n) = \frac{\mu}{\sigma_w^2 \pi} \left( \frac{\mu}{\xi} \right)^\mu \left( \frac{\mu}{\xi} + \frac{|w_n|^2}{\sigma_w^2} \right)^{-(\mu+1)},
\end{equation}
where $\mu \in (1, +\infty)$ is the shape parameter controlling the non-Gaussianity of $w_n$ and the scale parameter is defined as $\xi = \frac{\mu}{\sigma_w^2 (\mu - 1)}$.
In the simulations, the parameters used are $p=6$ for the order of the AR process, $\mu = 2$, $\sigma_w^2 = 1$ and the coefficient vector $\boldsymbol{\rho}$ is defined as:
\begin{align}
    \boldsymbol{\rho} &= [0.5e^{-j2\pi \cdot 0.4}, 0.6e^{-j2\pi \cdot 0.2}, 0.7e^{-j2\pi \cdot 0}, \notag \\
    &\quad 0.4e^{-j2\pi \cdot 0.1}, 0.5e^{-j2\pi \cdot 0.3}, 0.6e^{-j2\pi \cdot 0.35}]^T.
\end{align}
The number of spatial channels $N = N_T N_R = 10^4$, the number of angle bins $L = N_T = 100$, the total power $P_T = 1$ and the probability of false alarm $P_{FA} = 10^{-4}$.\\
The table \ref{tab:pomcp_parameters} contains the parameter settings of the POMCP algorithm.

\begin{table}[ht]
    \centering
    \caption{POMCP parameters}
    \begin{tabular}{|c|c|c|c|}
        \hline
        Parameter & Value & Parameter & Value \\
        \hline
        Number of simulations $N_{\text{sim}}$ & $10^{4}$ & Discount factor $\gamma$ & 0.8 \\
        Number of particles $N_p$  & $10^{4}$ & Time steps & 100 \\
        UCB1 parameter $c$ & $\sqrt{2}$ & Tree Depth & 2 \\
        \hline
    \end{tabular}
    \label{tab:pomcp_parameters}
\end{table}
A step size satisfying the condition \eqref{stepsize_choice} is chosen as $\beta_l = \sqrt{3} \widehat{\sigma}_l$ for all $l \in {1, \dots, L}$. Results are averaged over 250 Monte Carlo simulations. \\
To establish a lower bound for performance evaluation, we define an Oracle algorithm. The Oracle is an idealized algorithm that knows the future angle bin containing the target (but not its exact coordinates or velocity) and always takes the optimal action based on this knowledge. It receives an observation and builds a belief state. In essence, the oracle is the equivalent of a flawless POMCP. \\
The POMCP, particle filter, and Oracle are compared for the tracking task using the root-mean-square error (RMSE) for coordinate and velocity estimation. For the detection task, these three algorithms, along with the SARSA algorithm and the orthogonal waveform, are evaluated based on the probability of detection $P_D$.

\subsection{Study Case 1: Slow target}
\begin{figure}
\centering
\includegraphics[scale = 0.6]{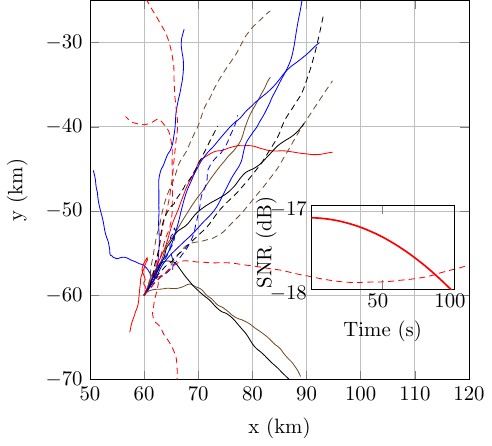}   
\caption{Study case 1: Potential trajectories with $\mathbf{s}_0 = (60\text{km}, 0.2\text{km/s}, -60\text{km}, 0.2\text{km/s})^T$ and noise $\sigma_s = 0.03$. The inner figure represents the average of SNR trajectories.}
\label{fig:trajectories_with_snr_slow}
\end{figure}

In this scenario, one target is considered and initialized with $\mathbf{s}_0 = (60\text{km}, 0.2\text{km/s}, -60\text{km}, 0.2\text{km/s})^T$, with noise $\sigma_s = 0.03$. Figure \ref{fig:trajectories_with_snr_slow} shows potential trajectories and the average SNR trajectories, which start at $-17\text{dB}$ and decrease to $-18\text{dB}$ after 100-time steps, i.e., most SNR trajectories decrease.

\begin{figure}
\centering
\includegraphics[scale = 0.6]{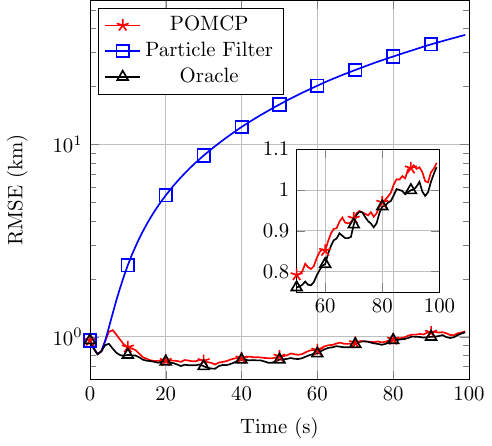} 
\caption{Study case 1: RMSE between the estimated and true coordinates of the target for each algorithm.}
\label{fig:rmse_coordinates}
\end{figure}

\begin{figure}
\centering
\includegraphics[scale = 0.6]{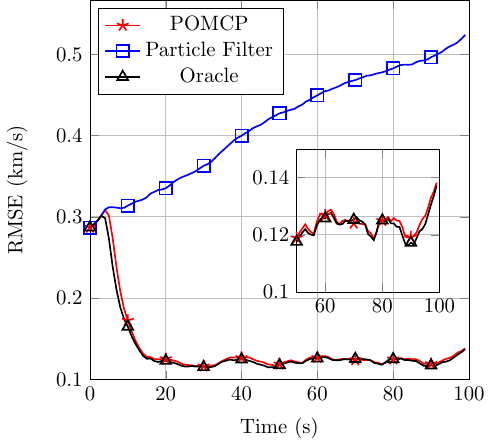}
\caption{Study case 1: RMSE between the estimated and true velocities of the target for each algorithm.}
\label{fig:rmse_velocities}
\end{figure}

\begin{figure}
\centering
\includegraphics[scale = 0.6]{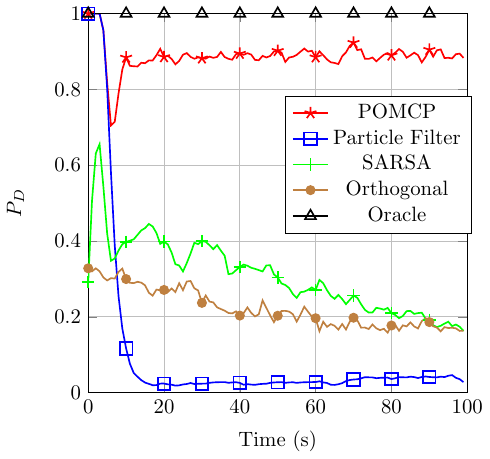}
\caption{Study case 1: Probability of detecting a moving target for different approaches.}
\label{fig:pd}
\end{figure}

Figure \ref{fig:rmse_coordinates} shows an increasing RMSE for the particle filter, which can be explained by the fact that the target becomes lost early on, which is confirmed in Figure \ref{fig:pd}. However, the POMCP estimates stay close to Oracle ones.\\
Figure \ref{fig:rmse_velocities} shows a bad velocity estimation at the start of the simulation but improves as the POMCP gathers more observations, which leads to the filtration of particles that do not represent the hidden state.\\

As shown in Figure \ref{fig:pd}, the particle filter loses the target as it moves, eventually leading to bad position and velocity estimations.
It also shows a robust performance of the POMCP in keeping the probability of detection high, significantly outperforming the SARSA and the particle filter, which drops below $0.2$. Given that the SNR is low, the orthogonal waveform does not have a high probability of detection.

\subsection{Study Case 2: Fast target}
\begin{figure}
\centering
\includegraphics[scale = 0.6]{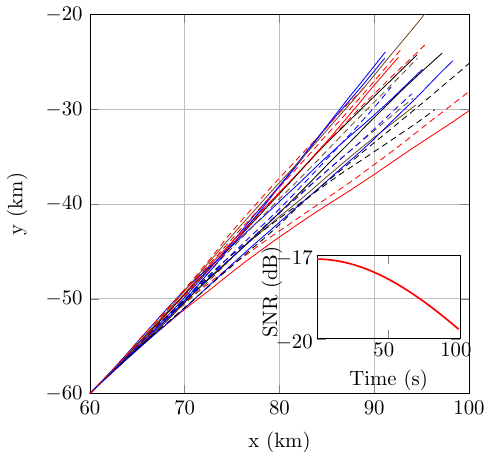}
\caption{Study case 2: Potential trajectories with $\mathbf{s}_0 = (60\text{km}, 0.35\text{km/s}, -60\text{km}, 0.35\text{km/s})^T$ and noise $\sigma_s = 0.005$. The inner figure represents the average of SNR trajectories.}
\label{fig:trajectories_far}
\end{figure}

In this scenario, one target is considered and initialized with $\mathbf{s}_0 = (60 \text{km}, 0.35 \text{km/s}, -60 \text{km}, 0.35 \text{km/s})^T$ and $\sigma_s=0.005$. \\
Figure \ref{fig:trajectories_far} shows that the target has linear trajectories given its high speed. In the inner plot, the SNR trajectories, on average, start from $-17\text{dB}$ and decrease to $-20\text{dB}$ after 100-time steps.

\begin{figure}
\centering
\includegraphics[scale = 0.6]{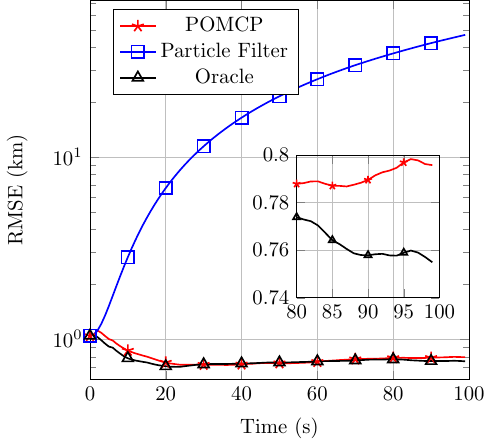}
\caption{Study case 2: RMSE between the estimated and true coordinates of the target for each algorithm. The inset highlights the close behavior of the POMCP and Oracle curves.}
\label{fig:rmse_coordinates_far}
\end{figure}

\begin{figure}
\centering
\includegraphics[scale = 0.6]{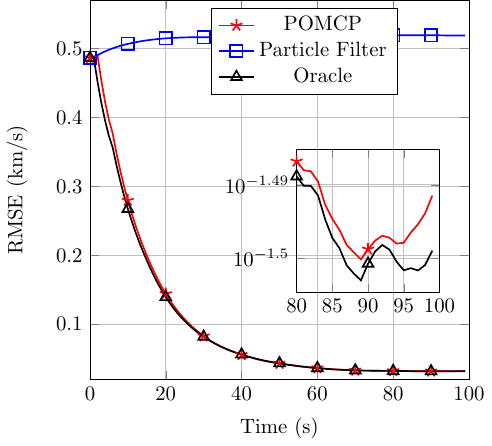}
\caption{Study case 2: RMSE between the estimated and true velocities of the target for each algorithm.}
\label{fig:rmse_velocities_far}
\end{figure}

\begin{figure}
\centering
\includegraphics[scale = 0.6]{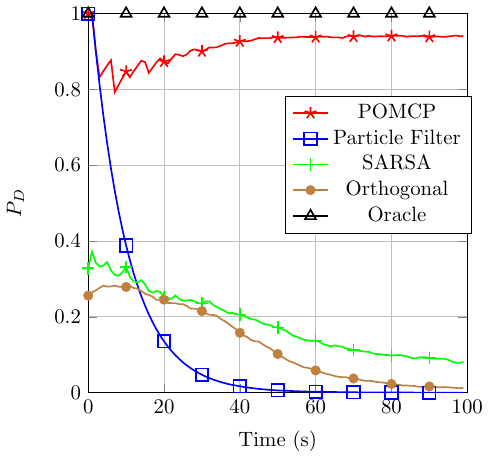}
\caption{Study case 2: Probability of detecting a moving target for different approaches.}
\label{fig:pd_far}
\end{figure}

Figures \ref{fig:pd_far}, \ref{fig:rmse_velocities_far} and \ref{fig:rmse_coordinates_far} present similar patterns to the previous study case. The RMSE values for POMCP estimates demonstrate smaller deviations from the Oracle ones than the particle filter estimates. The detection probabilities in Figure \ref{fig:pd_far} indicate that target detection with the POMCP remains feasible under these conditions.

Given that the SNR in this case is much lower than the previous one, it is normal that the probability of detection with the orthogonal waveform is low. The POMCP manages to detect a fast moving target and maintains a probability of detection above $0.8$.

\subsection{Tuning the depth of the tree in POMCP} 
In the previous study cases, the depth of the tree was chosen to be equal to 2 for two main reasons:
\begin{itemize} 
    \item \textbf{Short-term planning}: A shallow tree depth (2–3 steps) is sufficient to optimize actions for the short-term goal of estimating the hidden state at each time step. 
    \item \textbf{Cancel the generator's bias:} Increasing the tree depth would cause the tree's beliefs to be dominated by particles generated by the proposed generator $\mathcal{G}(\mathbf{s}_t, a_t)$, rather than particles obtained from real-world observations.
\end{itemize}
We compare the results obtained when the tree depth is increased from 2 to 5, while keeping the number of particles the same.
\begin{figure}[h]
    \centering
    \begin{subfigure}[b]{0.2\textwidth}
        \centering
        \includegraphics[scale=0.7]{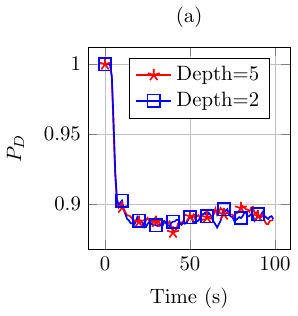}
        \label{fig:sub1}
    \end{subfigure}
    \hfill 
    \begin{subfigure}[b]{0.2\textwidth}
        \centering
        \includegraphics[scale=0.7]{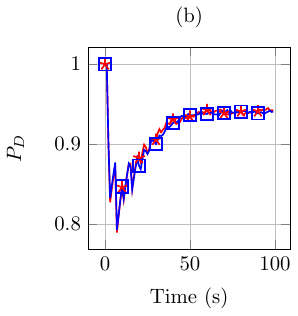}
        \label{fig:sub2}
    \end{subfigure}
    
    \vspace{2em} 
    \begin{subfigure}[b]{0.2\textwidth}
        \centering
        \includegraphics[scale=0.7]{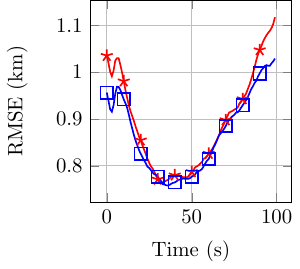}
        \label{fig:sub3}
    \end{subfigure}
    \hfill
    \begin{subfigure}[b]{0.2\textwidth}
        \centering
        \includegraphics[scale=0.7]{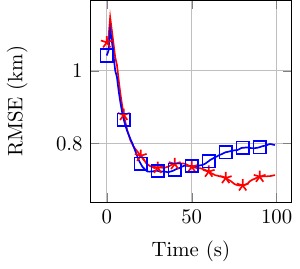}
        \label{fig:sub4}
    \end{subfigure}
    
    \vspace{2em} 
    \begin{subfigure}[b]{0.2\textwidth}
        \centering
        \includegraphics[scale=0.7]{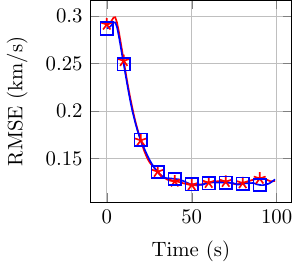}
        \label{fig:sub5}
    \end{subfigure}
    \hfill
    \begin{subfigure}[b]{0.2\textwidth}
        \centering
        \includegraphics[scale=0.7]{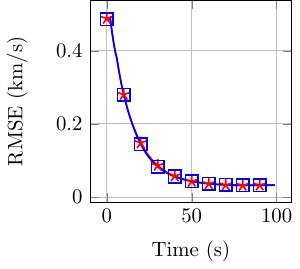}
        \label{fig:sub6}
    \end{subfigure}
    
    \caption{Results obtained by increasing tree depth from 2 to 5.
    (a) Study case 1, (b) Study case 2.}
    \label{depth_5}
\end{figure}
Figure \ref{depth_5} demonstrates the effects of tree depth. Increasing the depth from 2 to 5 slightly improved Cartesian coordinate estimation in study case 2. In the other study case, a deeper tree caused a slight decline in position estimation accuracy, while the $P_D$ and velocity estimation remained unaffected in both study cases. \\
As an illustration, in the simulations, smaller tree depths are more suitable for the highly dynamic target (study case 1). In contrast, stable trajectories in study case 2 may have benefited from higher tree depths. 

\subsection{Actions taken by the POMCP}
\begin{figure}[h]
    \centering
    \begin{subfigure}[b]{0.2\textwidth}
        \centering
        \includegraphics[scale=0.7]{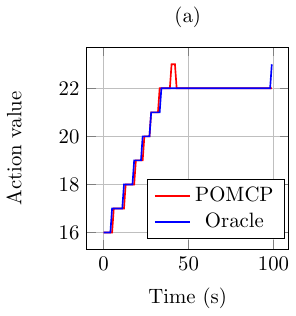}
        \label{fig:sub1_a}
    \end{subfigure}
    \hfill 
    \begin{subfigure}[b]{0.2\textwidth}
        \centering
        \includegraphics[scale=0.7]{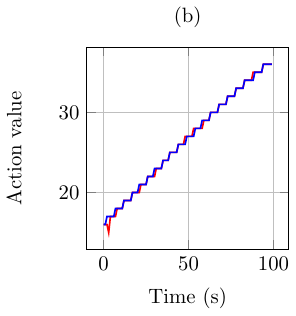}
        \label{fig:sub2_a}
    \end{subfigure}
    
    \vspace{2em} 
    \begin{subfigure}[b]{0.2\textwidth}
        \centering
        \includegraphics[scale=0.7]{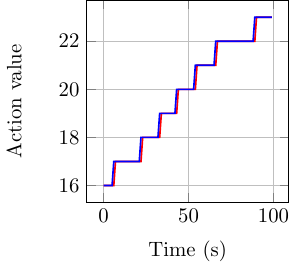}
        \label{fig:sub3_a}
    \end{subfigure}
    \hfill
    \begin{subfigure}[b]{0.2\textwidth}
        \centering
        \includegraphics[scale=0.7]{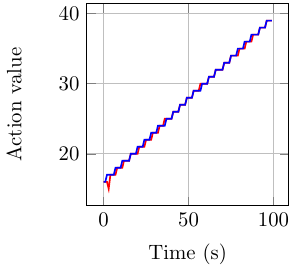}
        \label{fig:sub4_a}
    \end{subfigure}
    
    \vspace{2em} 
    \begin{subfigure}[b]{0.2\textwidth}
        \centering
        \includegraphics[scale=0.7]{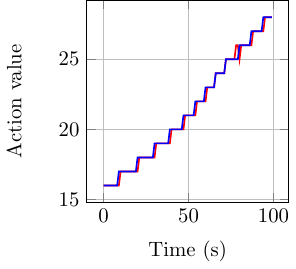}
        \label{fig:sub5_a}
    \end{subfigure}
    \hfill
    \begin{subfigure}[b]{0.2\textwidth}
        \centering
        \includegraphics[scale=0.7]{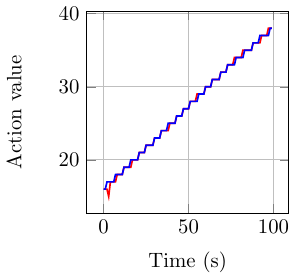}
        \label{fig:sub6_a}
    \end{subfigure}
    
    \caption{Examples of actions chosen by the POMCP compared to the optimal actions. (a) Study case 1,
    (b) Study case 2.}
    \label{actions_time}
\end{figure}

This section aims to demonstrate the actions taken by the POMCP and how they evolve over time. Given that the particle filter performed badly, this section will only compare the POMCP actions against the optimal actions. In essence, the POMCP has access to $L=100$ actions, which is to select the potential angle bin $\theta_l$ where the target could be, and after that, the waveform matrix is computed given the chosen angle bin. This suggests that optimal actions depend only on the trajectory of the target. Figure \ref{actions_time} confirms this suggestion as the six examples show that the optimal action the radar should choose changes over time.

\subsection{Other hyperparameters}
The algorithm involves numerous hyperparameters, some influencing the radar's number of virtual channels $N$, while others control the POMCP algorithm's behavior. Additional simulations were conducted by varying the $P_{FA}$ and the clutter disturbance distribution (using a Generalised Gaussian for the innovations). However, these variations yielded results consistent with our previous findings, with the POMCP successfully tracking and detecting the target. Given the similarity of these outcomes to the presented results, we opted not to include these additional simulations in the paper.

\section{Conclusion}
This work presents an original POMCP framework for the joint detection and tracking of a moving target in disturbance-agnostic scenarios for Massive MIMO radar systems. Our results show that POMCP maintains reliable detection probabilities even under unknown disturbances, successfully tracking targets across varying velocities. This study validates the POMCP as a practical solution for real-world radar applications. Future work will focus on developing a comprehensive understanding of the impact of hyperparameters on algorithm performance. Additionally, efforts will be directed towards extending this framework to multi-target scenarios. This extension would involve adapting the POMDP model to handle multiple state spaces and exploring more efficient particle sampling strategies to manage the increased computational complexity while ensuring real-time performance.

\label{sec:conclusion}


\bibliographystyle{ieeetr}
\bibliography{bib}

%








\end{document}